\documentclass{article}

\usepackage{amsmath}
\usepackage{array}
\usepackage{booktabs}
\usepackage{float}
\usepackage{graphicx}
\usepackage{multirow}
\usepackage{xcolor}
\usepackage{marvosym}

\usepackage[preprint]{corl_2026} 

\title{iMaC: Translating Actions into Motion and Contact Images for Embodied World Models}

\author{
Zhenyu Wu$^{1*}$,
  Xiuwei Xu$^{2*}$,
  Yukun Zhou$^{3}$,
  Yifan Li$^{3}$,
  Qiuping Deng$^{3}$,
  Xiaofeng Wang$^{3}$, \\
  \textbf{Zheng Zhu}$^{3}$,
  \textbf{Bingyao Yu}$^{2}$,
  \textbf{Ziwei Wang}$^{4}$,
  \textbf{Jiwen Lu}$^{2}$,
  \textbf{Haibin Yan}$^{1}$\textsuperscript{\Letter}\\
  $^{1}$Beijing University of Posts and Telecommunications\\
  $^{2}$Tsinghua University \qquad
  $^{3}$GigaAI \qquad
  $^{4}$Nanyang Technological University
}

\begin{document}
\maketitle
\begin{center}
    \vspace{-0.5em}
    \normalsize\textbf{Project page:} \url{https://imac-wm.github.io/}
    \vspace{0.3em}
\end{center}


\begin{abstract}
    Embodied world models promise to serve as real-world simulators for robot policy evaluation and closed-loop rollout, but their reliability depends on how precisely they condition future video prediction on robot actions. Existing action-conditioned video models often encode future actions as compact vectors and inject them through learned conditioning modules, leaving the model to infer fine-grained spatial consequences indirectly. This abstraction is limiting for real manipulation, where centimeter-level action differences can determine contact, object motion, and task outcome. Toward more spatially explicit action conditioning, we present \textbf{i}mages of \textbf{M}otion \textbf{a}nd \textbf{C}ontact (iMaC), an embodied world model that converts future actions into image-like controls to guide video generation with precise robot appearance and robot-scene spatial relations. iMaC first uses the robot URDF and forward kinematics to render future robot-observation control videos (i.e., motion images) from future joint actions. It also predicts depth as an auxiliary signal to strengthen spatial understanding, and uses the resulting 3D pointclouds to build two-stream geometry controls (i.e., contact images) between the current scene and future robot. These controls describe both the future robot observations and the spatial interactions that drive scene dynamics. To enhance long-horizon manipulation, iMaC further leverages a training-time rollout strategy to support minute-level generation and reduce exposure bias across generated chunks. Experiments on eight challenging long-horizon real-robot manipulation tasks show that iMaC can evaluate the relative performance of different policy checkpoints, with world-model success estimates strongly positively correlated with real-world policy performance.

\end{abstract}


\section{Introduction}
World models~\citep{hafner2019dream,hafner2020mastering,hafner2023mastering,yang2023unisim,zhou2024robodreamer,garrido2026learning} have long been viewed as a foundation for planning and control: an agent can choose actions by predicting their consequences before executing them in the real world. Recent progress in video generation has renewed this idea for robotics, where future states can be represented directly as visual observations rather than manually engineered simulator states. Such embodied world models are especially attractive for robot policy evaluation~\citep{guo2025ctrlworld,team2025veoeval,wang2026interactive}. Real-world evaluation is slow, expensive, difficult to reproduce across policy checkpoints, and often unsafe for rare failure cases. In contrast, a learned real-world simulator can roll out a policy in generated observations, providing a scalable way to compare policies, analyze failures, and support closed-loop improvement without building task-specific physical simulators or collecting large numbers of hardware trials. For this use case, the model must also sustain long closed-loop rollouts: after each generated chunk, its own predicted observation becomes the next reference, so small visual or geometric errors can accumulate over time.

The usefulness of a learned real-world simulator depends critically on whether the world model is truly responsive to the robot's actions. A policy evaluator must not only produce realistic videos; it must predict how different future actions change contact, object motion, and task outcome. This requirement is particularly stringent in manipulation, where a few centimeters can decide whether a gripper touches an object, misses it, pushes it into a different pose, or causes an entirely different downstream trajectory. Existing robotic world models have made important progress toward controllable video prediction, but many of them still encode actions or proxy actions as compact vectors~\citep{guo2025ctrlworld,gao2026dreamdojo,wang2026interactive} and inject them through mechanisms such as cross-attention~\citep{vaswani2017attention}, AdaLN~\citep{peebles2023scalable}, FiLM~\citep{perez2018film}, or related learned conditioning modules. Such designs are convenient for adapting large video generators, yet the model must infer precise spatial consequences from an abstract signal. Other works attempt to express actions more explicitly: EVAC~\citep{jiang2025enerverse} and ABot-PhysWorld~\citep{chen2026abotphysworld} use projected spheres or projection-based action maps, while Action Images~\citep{zhen2026action} uses Gaussian mixture maps to represent actions. These representations make actions more visible to the video model, but they remain primarily action visualizations. They do not directly control the future robot body state in the generated frames, nor do they explicitly describe the interaction geometry between the future robot and the current scene. Spatially precise action representation that controls both future robot motion and contact-relevant robot-scene geometry therefore remains under exploration.

In this paper, we present \textbf{i}mages of \textbf{M}otion \textbf{a}nd \textbf{C}ontact (iMaC), an embodied world model that translates future actions into dense image-like controls for future-video prediction. Given an initial multi-view RGB observation and a future action sequence, iMaC predicts the future video while using the action sequence to specify not only where the robot will move, but also how that motion relates geometrically to the observed scene. It first converts future joint actions into \emph{motion images}: rendered robot-observation control videos produced by applying the robot URDF and forward kinematics to obtain future robot configurations and rendering the future robot body from the camera views. These controls specify the robot's future visual state directly, reducing the burden on the model to hallucinate robot motion from a compact action vector. To capture how this motion may affect the scene, iMaC further constructs \emph{contact images}, two-stream geometry controls built from robot and scene pointclouds. One stream measures distances from the current scene to the future gripper, while the other measures distances from the future robot to the current scene, encoding contact-relevant spatial relations between action-induced robot motion and the observed environment. These motion and contact images are injected as video controls, preserving the scalability of image-to-video modeling while making action conditioning spatially explicit. To support long-horizon evaluation, we further propose training-time rollouts in which generated chunks provide the next reference observation, reducing the train-test mismatch that arises during closed-loop generation. We conduct experiments on eight challenging long-horizon real-robot manipulation tasks and show that iMaC can rank different policies with different checkpoints by performance, with world-model evaluation scores strongly positively correlated with real-world success rates.

\section{Related Work}
    \textbf{Video Generation Models for Robotics:}
Video generation models are increasingly used in robotics for offline data generation, including cross-embodiment demonstration transfer, diverse visual composition, missing-view synthesis, human-robot demonstration alignment, and larger synthetic data engines built with 3D reconstruction or scene editing~\citep{liu2025robotransfer,tong2025fidelity,qian2025wristworld,li2025mimicdreamer,zhao2025real2edit2real,team2025gigaworld}. Another line uses video models as embodied world models for forecasting observations in planning, policy decoding, manipulation, training, and evaluation~\citep{yang2023unisim,du2023unipi,zhou2024robodreamer,jang2025dreamgen,guo2025ctrlworld,gao2026dreamdojo,wang2026interactive,kim2026cosmospolicy,jiang2025enerverse,xiao2025worldenv,chen2026abotphysworld}. These works show the value of scalable video priors, but manipulation rollout remains sensitive to action representation: compact or latent actions learn spatial consequences indirectly, while sparse projected maps expose limited geometry. iMaC addresses this bottleneck by translating future actions into URDF/FK-based motion images and two-stream pointcloud-based contact images.

\textbf{Evaluation for Robotic Policies:}
Reliable policy evaluation is a central bottleneck for robot learning: real-world trials are authoritative but expensive and difficult to reproduce across checkpoints or rare failure cases~\citep{kressgazit2024robot,atreya2025roboarena,zhou2025autoeval,wang2025roboeval}. Physics simulators, manipulation benchmarks, and real-to-sim or digital-twin systems improve repeatability and alignment~\citep{todorov2012mujoco,zhu2020robosuite,xiang2020sapien,liu2023libero,pumacay2024colosseum,li2024evaluating,torne2024reconciling,badithela2025reliable,zhang2025realtosim,meyer2024pegasus}, but they still require assets, tuned dynamics, and careful scene construction. Video world models offer a complementary in-silico evaluator, with recent work using action-conditioned rollouts to compare policies, test OOD or safety settings, and obtain scores correlated with real-world performance~\citep{guo2025ctrlworld,quevedo2025worldgym,team2025veoeval,wang2026interactive,majumdar2025predictive}. iMaC follows this direction while targeting the action sensitivity of long-horizon manipulation through explicit robot-motion and contact-geometry controls.

\section{Approach}
    \subsection{Problem Formulation and Overview}
\label{sec:problem}
\textbf{World model for policy evaluation:}
Let $o_t$ denote the robot RGB observation at time $t$. A robot policy $\pi$ maps $o_t$ and an optional language instruction $l$ to a future action sequence,
\begin{equation}
    a_{t:t+H-1} = \pi(o_t, l).
\end{equation}
Given the current observation and the policy-proposed actions, an action-conditioned world model predicts the future observation chunk
\begin{equation}
    \hat{o}_{t+1:t+H} = f_\theta(o_t, a_{t:t+H-1}, l).
\end{equation}
During policy evaluation, $\pi$ and $f_\theta$ form a closed loop: the policy acts on generated observations, and the world model predicts the visual consequences of those actions. This paper focuses on learning $f_\theta$, not improving $\pi$ itself, but the core objective is to make world-model rollouts reliable enough to compare policy checkpoints and estimate their real-world performance.

\textbf{IT2V world-model backbone:}
iMaC builds on a WAN2.2 image-to-video (IT2V) DiT~\citep{team2025wan}. In our implementation, $o_t$ contains one fixed head-camera view and two wrist-camera views, arranged as a single image mosaic so that multi-view prediction can still be handled by a single-image IT2V model. Rollout is generated chunk-wise. The first chunk uses the given initial image as the reference image; for later chunks, the last generated frame of the previous chunk becomes the next reference image. For each chunk, the reference image and target future video are encoded together by the WAN VAE encoder $\mathcal{E}$. Let $\mathbf{z}^{r}$ be the clean reference latent and $\mathbf{x}_1$ be the clean future-video latent. We sample $\mathbf{x}_0\sim\mathcal{N}(0,\mathbf{I})$, $\tau\sim\mathcal{U}(0,1)$, and noise only the future latent:
\begin{equation}
    \mathbf{x}_\tau=(1-\tau)\mathbf{x}_0+\tau\mathbf{x}_1 .
\end{equation}
iMaC constructs three action-derived control videos: motion images $\mathbf{C}^{m}$, scene-to-gripper contact images $\mathbf{C}^{s\rightarrow g}$, and robot-to-scene contact images $\mathbf{C}^{r\rightarrow s}$. After VAE encoding and control-specific patchification, these controls are added to the noised future tokens, while the reference tokens remain clean:
\begin{equation}
    \mathbf{h}_\tau
    =
    \left[
    P_v(\mathbf{z}^{r})\ ;\
    P_v(\mathbf{x}_\tau)
    + P_m(\mathcal{E}(\mathbf{C}^{m}))
    + P_{s\rightarrow g}(\mathcal{E}(\mathbf{C}^{s\rightarrow g}))
    + P_{r\rightarrow s}(\mathcal{E}(\mathbf{C}^{r\rightarrow s}))
    \right],
    \label{eq:dit_input}
\end{equation}
where $P_v$ is the WAN video patchify layer and $P_m$, $P_{s\rightarrow g}$, and $P_{r\rightarrow s}$ are control-specific patchify layers. The DiT predicts the flow only for future tokens, with objective
\begin{equation}
    \mathcal{L}_{fm}
    =
    \mathrm{E}_{\mathbf{x}_0,\mathbf{x}_1,\tau}
    \left[
    \left\|
    v_\theta(\mathbf{h}_\tau,\tau,l)
    -
    (\mathbf{x}_1-\mathbf{x}_0)
    \right\|_2^2
    \right],
    \label{eq: training}
\end{equation}
Besides RGB prediction, iMaC also constructs an auxiliary depth prediction branch, which provides geometric state for constructing pointcloud controls in subsequent chunks. Sec.~\ref{sec:motion_images} converts actions into URDF/FK-rendered motion images that specify future robot appearance, Sec.~\ref{sec:contact_images} builds two-stream contact images from robot and scene pointclouds to encode contact-relevant geometry, and Sec.~\ref{sec:training_rollout} describes training-time rollouts for long-horizon generation.

\begin{figure}[!t]
    \centering
    \includegraphics[width=\textwidth]{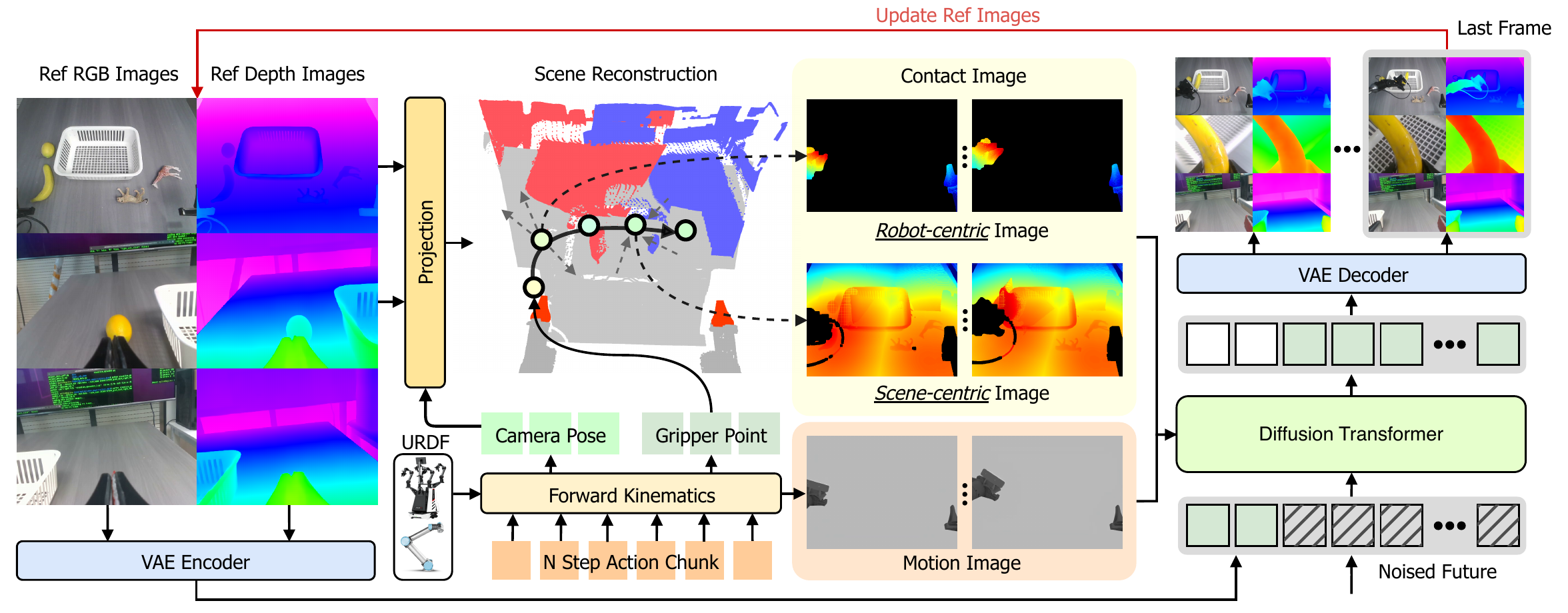}
    \caption{Overall iMaC pipeline. Given a reference observation and future actions, iMaC translates actions into motion images from robot kinematics and contact images from robot-scene geometry, injects these image-like controls into an IT2V world model, and rolls out future video chunks for policy evaluation.}
    \label{fig:imac_pipeline}
\end{figure}

\subsection{Motion Images from Robot Kinematics}
\label{sec:motion_images}

Predicting a future manipulation video can be decomposed conceptually into predicting the future robot appearance and predicting the future scene response. The scene response is governed by contact-rich physical dynamics and is difficult to infer from actions alone. In contrast, the robot part of the future video is largely determined by the commanded future action sequence and the robot kinematic model. This distinction is important for policy evaluation: if the generated gripper or arm deviates from the action actually proposed by the policy, the subsequent contact pattern can be wrong even when the rendered scene remains visually plausible. iMaC therefore avoids asking the world model to infer future robot motion only through a compact action embedding. Instead, it translates the future actions into dense robot-observation control videos, i.e., motion images.

Given the future joint action sequence $a_{t:t+H-1}$, iMaC applies the robot URDF and forward kinematics to obtain the corresponding future robot configurations. Let $\phi$ denote the robot controller's action-to-joint-state update, and let $\mathcal{K}_{\mathrm{URDF}}$ be forward kinematics defined by the robot URDF. For the $k$-th future step and camera view $v$, we construct
\begin{equation}
    \mathbf{q}_{t+k}
    =
    \phi(\mathbf{q}_t, a_{t:t+k-1}),\quad
    \mathcal{M}_{t+k}
    =
    \mathcal{K}_{\mathrm{URDF}}(\mathbf{q}_{t+k}),\quad
    \mathbf{C}^{m}_{t+k,v}
    =
    \mathcal{R}(\mathcal{M}_{t+k}; K_v, T^v_{t+k}),
\end{equation}
where $\mathbf{q}_{t+k}$ is the future joint state, $\mathcal{M}_{t+k}$ denotes the posed robot model, $\mathcal{R}$ is the renderer, and $(K_v,T^v_{t+k})$ are the camera intrinsics and extrinsics. We render the robot model from the same camera views used by the world model, including the fixed head camera and the two wrist cameras whose poses are obtained from forward kinematics. The rendered three-view robot observations are arranged in the same mosaic format as the predicted video, yielding a control video $\mathbf{C}^{m}_{t+1:t+H}$. These motion images specify the future robot body and gripper appearance directly in image space. Since the control branch is injected by latent-wise addition to the noised future-video tokens in Eq.~\ref{eq:dit_input}, the rendered robot observations provide strong pixel-level guidance for the part of the future video whose geometry is known from the robot model.

iMaC is used as a world model for evaluating policies on a specified robot platform. In this setting, access to the robot URDF is a natural requirement, analogous to knowing the hardware platform in real-world policy evaluation. The construction therefore does not require additional human annotation or task-specific labeling beyond the future action sequence already needed for action-conditioned rollout.

\subsection{Contact Images from Robot-Scene Geometry}
\label{sec:contact_images}

\textbf{Taming World Model for RGB-D Prediction:}
Beyond RGB prediction, iMaC predicts depth to improve the world model's spatial understanding and to provide geometry for subsequent contact-control construction. The initial reference depth is estimated from the three RGB views and their camera poses, where the wrist-camera poses are obtained from URDF and forward kinematics, using Depth Anything 3 (DA3)~\citep{lin2025depth}. For later chunks, depth is predicted by the world model together with RGB, so the generated final frame provides both the next visual reference and geometric state.

To keep depth compatible with the image-to-video backbone, we encode each depth map as a colorized image following VisionBanana~\citep{gabeur2026image}. The model input and output are organized as a six-panel mosaic: the first row contains the three RGB views, and the second row contains the corresponding colorized depth views. RGB and depth therefore pass through the same VAE encoder $\mathcal{E}$. Since the future-video latent now has this six-panel layout, each three-view control video is vertically duplicated before control injection, so that the control tokens are spatially aligned with both the RGB and depth rows. We use this dimension-matched version of the controls in Eq.~\ref{eq:dit_input}, and the injection remains the same latent-wise addition.

\textbf{Two-stream Geometry Controls:}
Given the current predicted depth and the future actions, iMaC builds two contact-image streams following a bidirectional robot-scene distance construction. We first remove the robot from the current depth using the rendered robot mask at the reference step, then lift the remaining pixels from all views into a scene pointcloud $\mathbf{P}^{s}_{t}$. Future full-robot pointclouds $\mathbf{P}^{r}_{t+k}$ and gripper pointclouds $\mathbf{P}^{g}_{t+k}\subset\mathbf{P}^{r}_{t+k}$ are obtained from the URDF/FK-predicted robot configurations.

The first stream is robot-to-scene: each future robot point stores its nearest distance to the current scene,
\begin{equation}
    d^{r\rightarrow s}_{t+k}(\mathbf{r})
    =
    \min_{\mathbf{p}\in\mathbf{P}^{s}_{t}}
    \left\|
    \mathbf{r}-\mathbf{p}
    \right\|_2,
    \quad
    \mathbf{r}\in\mathbf{P}^{r}_{t+k}.
    \label{eq:robot_to_scene}
\end{equation}
These distances are projected with the future robot pose into the robot render mask and densified inside the mask, producing the robot-centric contact images $\mathbf{C}^{r\rightarrow s}_{t+1:t+H}$. The second stream is scene-to-gripper: each current scene point stores its nearest distance to the future gripper,
\begin{equation}
    d^{s\rightarrow g}_{t+k}(\mathbf{p})
    =
    \min_{\mathbf{g}\in\mathbf{P}^{g}_{t+k}}
    \left\|
    \mathbf{p}-\mathbf{g}
    \right\|_2,
    \quad
    \mathbf{p}\in\mathbf{P}^{s}_{t}.
    \label{eq:scene_to_gripper}
\end{equation}
These distances are projected back to the current scene pixels and densified inside the scene mask, producing the scene-centric contact images $\mathbf{C}^{s\rightarrow g}_{t+1:t+H}$. Both distance videos are colorized with a fixed heatmap after sequence-level distance normalization. Together, the two streams tell the world model where the future robot body approaches the scene and which current scene regions are close to the future gripper, providing contact-relevant spatial guidance beyond the rendered motion images.

\subsection{Training-time Rollout for Long Video Generation}
\label{sec:training_rollout}
Long-horizon policy evaluation requires the world model to operate on its own generated observations. At inference time, iMaC predicts a future chunk, takes the final generated RGB-D-style frame as the next reference, constructs new motion and contact images from the next action chunk and the predicted depth, and repeats this process. If training always conditions on ground-truth reference frames, the model only learns under clean contexts, while closed-loop evaluation conditions on imperfect generated contexts. This train-test mismatch is a form of exposure bias and can cause visual, depth, and contact-control errors to accumulate across chunks.

To reduce this mismatch, iMaC performs training-time rollout over multiple consecutive chunks. Given a training sequence, we split it into $R$ chunks of length $H$. For the first chunk, the reference is the ground-truth initial RGB-D-style observation. For chunk $r$, we train the model with the usual flow-matching objective in Eq.~\ref{eq: training} using the corresponding ground-truth future RGB and depth latents as targets and the action-derived controls for that chunk. Let $\mathbf{x}^{(r)}_1$ denote the clean RGB future latent and let $\mathbf{d}^{(r)}_1$ denote the clean depth future latent. We sample noise and $\tau$ as in Sec.~\ref{sec:problem}, form noisy latents $\mathbf{x}^{(r)}_\tau$ and $\mathbf{d}^{(r)}_\tau$, and obtain one-step clean estimates from the predicted flows,
\begin{align}
    \mathbf{x}^{(r)}_\tau
    &=
    (1-\tau)\mathbf{x}^{(r)}_0
    +
    \tau\mathbf{x}^{(r)}_1,
    \quad
    \mathbf{d}^{(r)}_\tau
    =
    (1-\tau)\mathbf{d}^{(r)}_0
    +
    \tau\mathbf{d}^{(r)}_1, \nonumber\\
    \hat{\mathbf{x}}^{(r)}_1
    &=
    \mathbf{x}^{(r)}_\tau
    +
    (1-\tau)\,
    v^x_\theta(\cdot),
    \quad
    \hat{\mathbf{d}}^{(r)}_1
    =
    \mathbf{d}^{(r)}_\tau
    +
    (1-\tau)\,
    v^d_\theta(\cdot).
\end{align}
where $v^x_\theta$ and $v^d_\theta$ are the RGB and depth flow predictions. The final frames decoded from $\hat{\mathbf{x}}^{(r)}_1$ and $\hat{\mathbf{d}}^{(r)}_1$ are detached and used as the reference RGB and depth for chunk $r+1$. Thus, later chunks are trained under generated references while still receiving paired supervision from the recorded video-action sequence. The depth reference is updated in the same way as RGB, so the model also learns to construct subsequent geometric state from its own depth predictions rather than from ground-truth.

Training under self-generated context is also studied in Self-Forcing~\citep{huang2026self}, but the objective and setting are different. Self-Forcing targets open-ended autoregressive video diffusion, where a text prompt can correspond to many plausible videos and a self-generated sample has no unique paired target; it therefore relies on distribution-matching objectives such as DMD~\citep{yin2024one,yin2024improved}, SiD~\citep{zhou2024score,zhou2025adversarial}, or GAN losses~\citep{goodfellow2014generative}. iMaC instead studies action-conditioned robot world modeling. Given the reference observation and future robot actions, each chunk in the recorded manipulation sequence provides aligned RGB-D supervision, even when the reference for that chunk is generated by the model. We therefore keep the standard flow-matching/MSE losses for RGB and depth. Operationally, iMaC also uses a simpler rollout update: rather than unrolling a full autoregressive diffusion sampler with rolling KV cache, it obtains the next reference by a one-step flow clean estimate, VAE-decodes the predicted RGB-D latents, detaches the final frame, and recomputes the next chunk's action-derived controls from the updated RGB-D reference.

\section{Experiment}
    \subsection{Experimental Setup}
\label{sec:exp_setup}

\textbf{Tasks and data.}
We evaluate iMaC on eight real-world manipulation tasks that require contact-sensitive prediction over closed-loop rollouts.
Each task contains paired multi-view RGB videos and robot action trajectories collected from a mixture of teleoperation and policy rollouts, including both successful and failed executions.
Appendix~\ref{app:tasks} provides task descriptions and visualizations.
The observation contains one fixed head-camera view and two wrist-camera views; during training, iMaC additionally uses the corresponding depth-color targets described in Sec.~\ref{sec:contact_images}.
At test time, policies act from RGB observations, while depth is only used internally to construct pointcloud-based contact images during chunk-wise rollout.

\textbf{World-model evaluation protocol.}
For policy evaluation, a policy is deployed in the learned world model in closed loop: the policy predicts the next action chunk from the current generated observation, iMaC predicts the future video chunk, and the final generated frame becomes the next reference.
We evaluate two VLA policy families, $\pi_{0.5}$~\citep{black2025pi_} and GigaBrain-0.5~\citep{team2026gigabrain}, using three checkpoints from each model.
The three checkpoints for each policy family sample early, intermediate, and late stages of training.
This protocol evaluates whether world-model scores preserve performance differences both across policy families and across training stages within each family.
Each checkpoint is evaluated on the same initial task configurations in both the real world and the iMaC world model, with one or two evaluation groups according to task availability.
Each group contains 30 episodes, and repeated groups provide an estimate of evaluation repeatability for the same policy checkpoint.
We treat real-world performance as the reference measurement and test whether normalized world-model scores preserve the relative ranking of policies and checkpoints, which is the key property needed for model-based checkpoint selection.

\textbf{Baselines and metrics.}
We compare iMaC with action-conditioned world-model baselines that inject future actions through learned action embeddings or sparse action-image controls.
The main video-prediction metrics are computed between generated and ground-truth future videos under the same initial observations and action sequences; Table~\ref{tab:video_quality} reports task-averaged video-quality metrics.
For policy evaluation, we report the correlation between world-model and real-world scores for each task.
We ablate URDF/FK-rendered motion images, two-stream contact images, and the source of depth used for contact-image construction.

\textbf{Implementation details.}
We train iMaC in two stages.
The first stage trains a shared model on data from all eight tasks, and the second stage finetunes a task-specific model on each individual task; the final evaluation therefore uses one world model per task.
Because contact images depend on the quality of predicted depth used for pointcloud construction, the first stage uses only motion-image controls, and contact-image controls are introduced during the second-stage finetuning after the model produces clearer RGB-D predictions.
Training-time rollout is also warmed up: for the first 40 epochs, chunks use clean reference observations before enabling one-step generated references.

\subsection{Main Results}
\label{sec:main_results}

\begin{table*}[t]
    \centering
    \caption{
    Quantitative comparison of future-video prediction quality.
    We report mean and standard deviation across the eight real-world tasks.
    Lower is better for MSE, FID, and FVD; higher is better for PSNR and SSIM.
    }
    \label{tab:video_quality}
    \scriptsize
    \setlength{\tabcolsep}{6pt}
    \resizebox{\textwidth}{!}{
    \begin{tabular}{lccccc}
        \toprule
        Method
            & MSE $\downarrow$
            & FID $\downarrow$
            & PSNR $\uparrow$
            & SSIM $\uparrow$
            & FVD $\downarrow$ \\
        \midrule
        Ctrl-World~\citep{guo2025ctrlworld}
            & 0.030 $\pm$ 0.012
            & 48.64 $\pm$ 10.68
            & 16.22 $\pm$ 1.74
            & 0.730 $\pm$ 0.037
            & 591.47 $\pm$ 160.30 \\
        ABot-PhysWorld~\citep{chen2026abotphysworld}
            & 0.041 $\pm$ 0.017
            & 74.23 $\pm$ 22.50
            & 14.41 $\pm$ 1.62
            & 0.630 $\pm$ 0.071
            & 642.98 $\pm$ 105.27 \\
        iMaC w/o contact images
            & \textbf{0.028 $\pm$ 0.009}
            & 38.81 $\pm$ 9.89
            & 16.34 $\pm$ 1.39
            & 0.735 $\pm$ 0.039
            & 523.94 $\pm$ 156.84 \\
        \textbf{iMaC}
            & 0.028 $\pm$ 0.010
            & \textbf{36.96 $\pm$ 9.16}
            & \textbf{16.39 $\pm$ 1.41}
            & \textbf{0.735 $\pm$ 0.037}
            & \textbf{489.51 $\pm$ 92.65} \\
        \bottomrule
    \end{tabular}
    }
\end{table*}

\begin{figure*}[t]
    \centering
    \includegraphics[width=\textwidth]{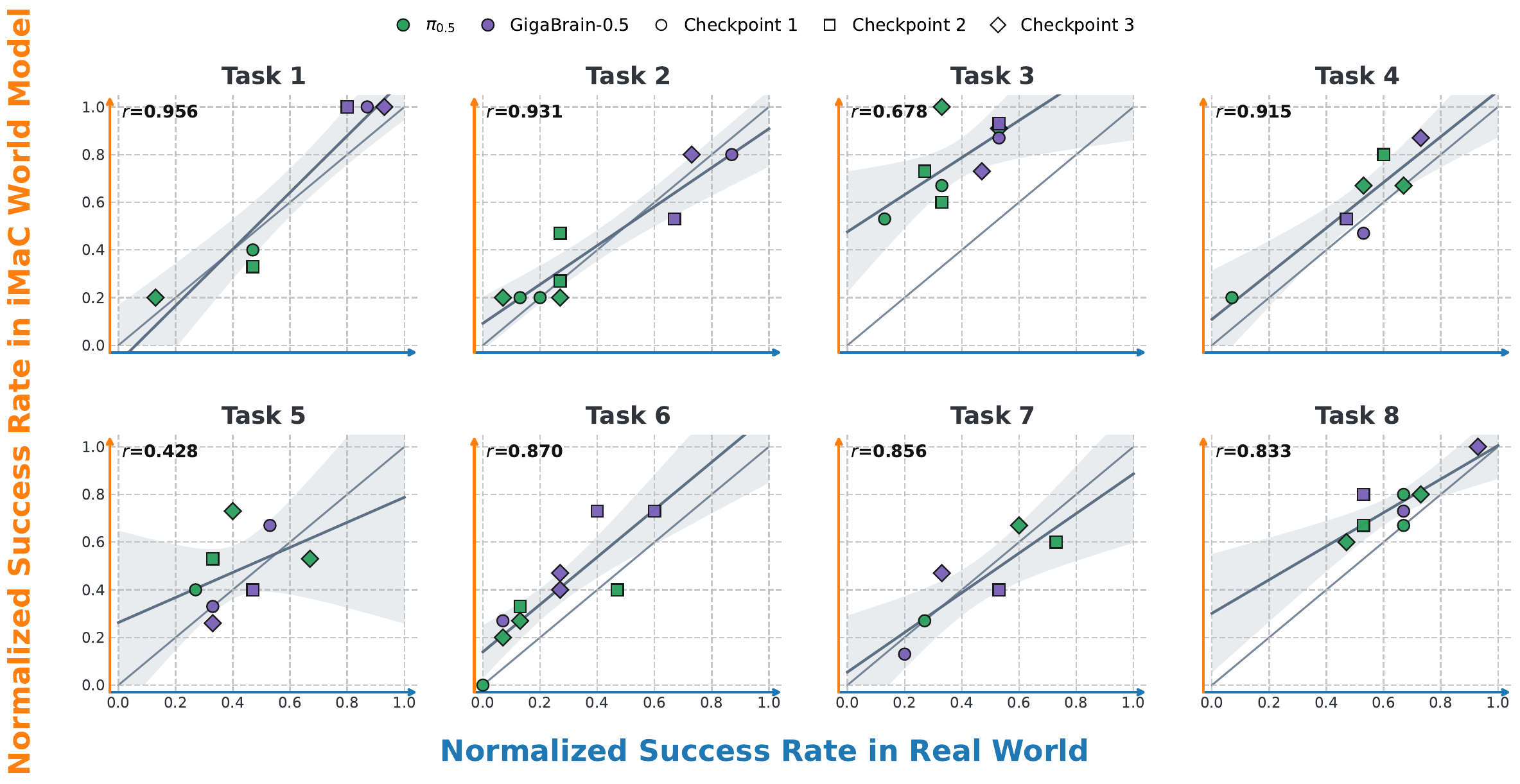}
    \caption{
    Correlation between normalized policy success rates measured in the iMaC world model and in the real world.
    Each subplot evaluates three checkpoints from $\pi_{0.5}$ and three checkpoints from GigaBrain-0.5 on matched initial configurations; repeated groups for the same checkpoint share the same color and marker.
    Most tasks show strong positive correlation, while Tasks 3 and 5 expose the missing-observation failure mode analyzed in Appendix~\ref{app:rollout_visualization} and Fig.~\ref{fig:app_rollouts_controls_failures}.
    }
    \label{fig:wm_real_correlation}
\end{figure*}

Table~\ref{tab:video_quality} evaluates future-video prediction before closed-loop policy evaluation.
iMaC obtains the best task-averaged FID, PSNR, SSIM, and FVD, while matching the best MSE within rounding, showing that motion images and contact images improve action-conditioned prediction quality by specifying future robot state and dense robot-scene distance cues.
Appendix~\ref{app:rollout_visualization} provides additional rollout and control-video visualizations.

Figure~\ref{fig:wm_real_correlation} evaluates iMaC as a closed-loop policy evaluator.
Across six of the eight tasks, world-model scores are strongly aligned with real-world performance, with per-task correlations between $0.833$ and $0.956$.
This indicates that iMaC usually preserves the relative ranking of policy families and training checkpoints, which is the key requirement for checkpoint selection before additional hardware evaluation.
The two lower-correlation tasks, Task 3 ($r=0.678$) and Task 5 ($r=0.428$), are not arbitrary outliers: both depend on height relations that are weakly observed by the available camera views.
In Task 3, the model must know whether the box ear has been lifted high enough to clear the side wall before entering the slot; in Task 5, it must know whether the dustpan entrance is flush with the tabletop or raised above the paper trash.
Appendix~\ref{app:rollout_visualization} analyzes these missing-observation cases.

\subsection{Ablation Study}
\label{sec:ablation}

\begin{figure*}[t]
    \centering
    \includegraphics[width=\textwidth]{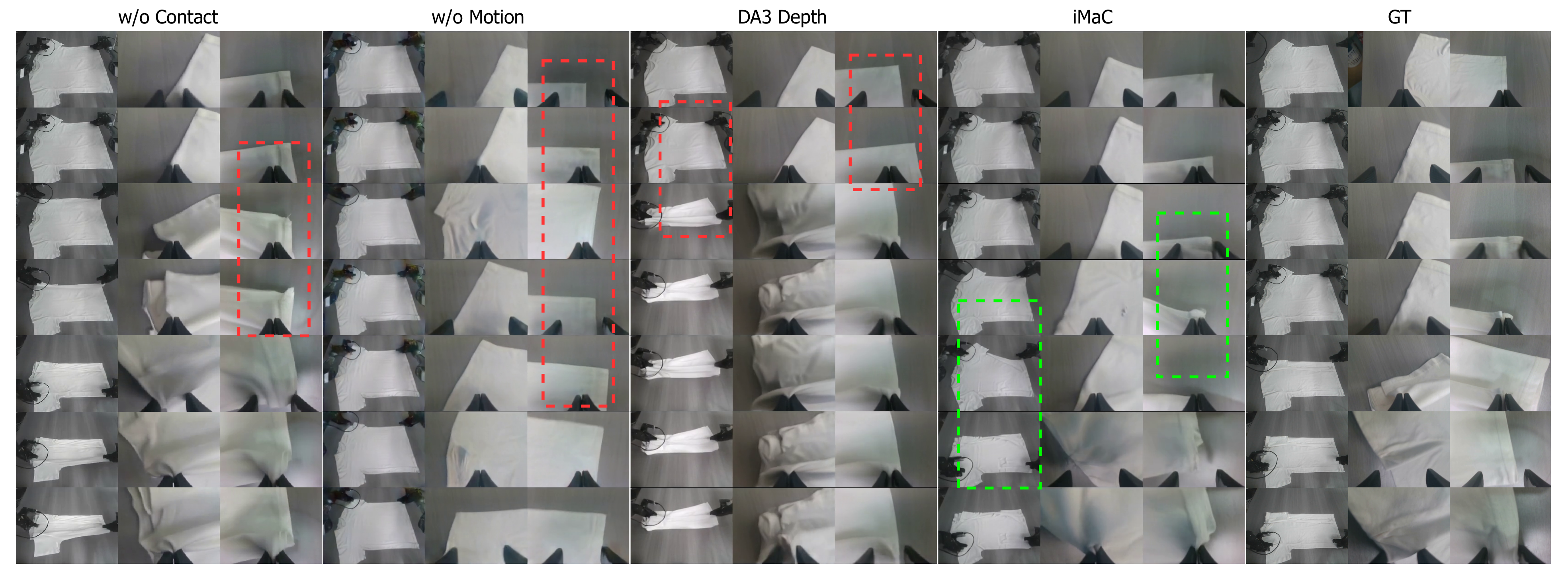}
    \caption{
    Ablation of contact/motion images and the depth source for contact-image construction.
    }
    \label{fig:ablation_video_quality}
\end{figure*}

\textbf{Motion and contact images.}
The first two columns isolate the two action-derived controls.
Without contact images, the world model lacks contact-aware guidance for action following: the gripper does not grasp the cloth, even though later frames still generate an interaction-like cloth motion.
Without motion images, the model lacks direct guidance about the future robot configuration; the gripper repeatedly attempts the motion but cannot produce the precise grasp.

\textbf{Depth source for contact images.}
The third and fourth columns keep motion and contact controls but compare how the depth used for contact-image construction is obtained.
Using DA3 depth partially improves action following because both controls are present, but its contact geometry is less consistent than iMaC's RGB-D world-model state: compared with the ground truth, the gripper still misses the cloth corner.
Additional qualitative rollout and control visualizations are provided in Appendix~\ref{app:rollout_visualization}.

\section{Limitation}
    iMaC relies on accurate 3D information to train depth prediction and to construct pointcloud-based contact images.
In the current system, depth supervision is estimated by Depth Anything 3 (DA3) from multi-view RGB observations and camera poses, which can introduce centimeter-level errors in manipulation scenes.
The two-stream contact images remain useful because they are heatmaps over distance fields, so the model can exploit coarse approaching and separating trends rather than exact metric contact at every pixel.
Higher-quality depth sensors or manipulation-adapted depth models should further improve contact timing, collision localization, and long-horizon rollout reliability.

\section{Concluding Remark}
    We presented iMaC, an embodied world model that translates future robot actions into motion images and contact images for spatially explicit action conditioning.
By combining URDF/FK-rendered robot controls, auxiliary depth prediction, two-stream pointcloud-based contact images, and training-time rollout, iMaC gives an image-to-video model direct guidance about future robot state and robot-scene geometry.
Across eight real-world manipulation tasks, iMaC is evaluated as a learned real-world simulator for ranking checkpoints of $\pi_{0.5}$ and GigaBrain-0.5, with world-model scores positively correlated with real-world performance.
iMaC is intended to complement, not eliminate, real-world evaluation: generated rollouts can help rank policies and reduce hardware trials, while final deployment decisions should still account for residual model bias and rare physical failures.


\bibliography{main}  

\appendix
\section*{Appendix}
\section{Task Suite}
\label{app:tasks}

\begin{figure}[H]
    \centering
    \includegraphics[width=\textwidth]{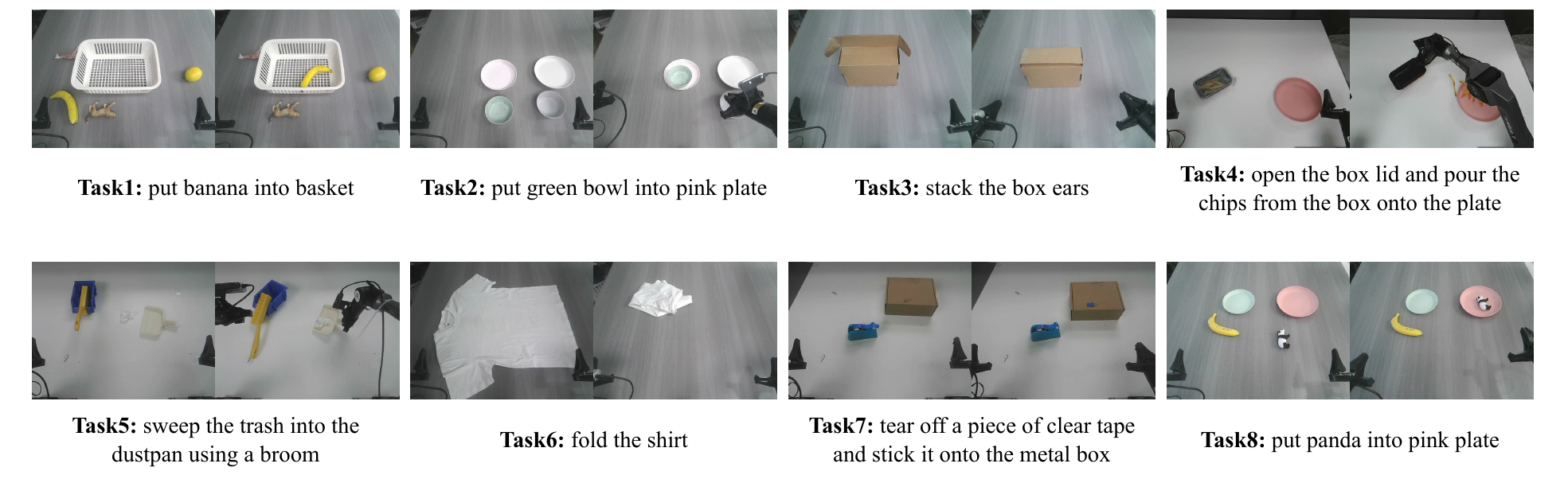}
    \caption{
    Visualization of the eight real-world manipulation tasks used for world-model-based policy evaluation.
    Each task is shown with an initial frame and a representative final frame from a successful rollout.
    Each task requires contact-sensitive prediction over closed-loop rollouts, so the world model must preserve both policy-dependent robot motion and the resulting scene changes.
    }
    \label{fig:app_tasks}
\end{figure}

This appendix provides additional materials for the eight real-world manipulation tasks used in Sec.~\ref{sec:exp_setup}.
Fig.~\ref{fig:app_tasks} visualizes the task suite with representative initial and final observations.
Across tasks, initial configurations vary in object placement, object pose, and robot approach state, while rollouts cover the full manipulation sequence from the initial observation to the task outcome.
These details complement the main paper's policy-evaluation results by clarifying what interaction outcomes the world model must predict during closed-loop rollout.

\paragraph{Task 1.}
The language instruction is ``put banana into basket.''
The scene contains a banana and a basket on the tabletop, with distractor objects possibly present in the workspace.
The policy must reach the banana, grasp or push it into a controllable pose, and place it inside the basket.
Success is achieved when the banana rests inside the basket at the end of the rollout.
This task tests whether the world model can predict object transport into a container while preserving the basket geometry and the banana's pose across multiple views.

\paragraph{Task 2.}
The language instruction is ``put green bowl into pink plate.''
The initial scene contains a green bowl, a pink plate, and additional bowls or plates that create visual ambiguity.
The policy must identify the green bowl, move it without disturbing the target plate excessively, and place it onto the pink plate.
Success is achieved when the green bowl is stably placed on the pink plate.
This task requires the generated rollout to preserve object identity, relative placement, and contact between two shallow objects.

\paragraph{Task 3.}
The language instruction is ``stack the box ears.''
The task starts from an open cardboard box whose side ears are outside the closing slot.
The policy must manipulate the ears so that they are folded and inserted into the slot rather than merely pushed against the box side.
Success is achieved when the box ears are stacked into the closed configuration.
This task is sensitive to small height and alignment differences: the world model must predict whether the ear clears the side wall and enters the slot, not only whether the visible box top appears closed.

\paragraph{Task 4.}
The language instruction is ``open the box lid and pour the chips from the box onto the plate.''
The initial scene contains a small chip box and a plate.
The policy must open the box, lift and tilt it, and pour the chips onto the plate.
Success is achieved when the chips are transferred from the box to the plate.
This task combines articulated object manipulation with granular object motion, requiring the world model to predict both the box pose and the downstream motion of the chips.

\paragraph{Task 5.}
The language instruction is ``sweep the trash into the dustpan using a broom.''
The scene contains scattered trash, a dustpan, and a broom.
The policy must control the broom to collect the trash and move it into the dustpan opening.
Success is achieved when the trash ends inside the dustpan.
This task stresses contact-rich pushing: small errors in broom pose, dustpan placement, or trash trajectory can change the final outcome.

\paragraph{Task 6.}
The language instruction is ``fold the shirt.''
The task starts from a spread-out shirt on the tabletop.
The policy must grasp and fold the cloth into a compact folded state.
Success is achieved when the shirt is folded into the desired final configuration.
This task requires the world model to forecast deformable object motion, including large nonrigid changes in cloth shape that are only partially constrained by the robot trajectory.

\paragraph{Task 7.}
The language instruction is ``tear off a piece of clear tape and stick it onto the metal box.''
The scene contains a tape dispenser, clear tape, and a metal box.
The policy must pull a piece of tape from the dispenser, tear it off, move it to the box, and press it onto the box surface.
Success is achieved when a piece of clear tape is attached to the metal box.
This task is visually challenging because the tape is small and transparent, so the world model must infer subtle contact and attachment states from limited visual evidence.

\paragraph{Task 8.}
The language instruction is ``put panda into pink plate.''
The scene contains a panda toy, a pink plate, and distractor objects such as a banana or another plate.
The policy must localize the panda toy, move it to the target plate, and release it inside the plate boundary.
Success is achieved when the panda rests on the pink plate at the end of the rollout.
This task evaluates whether the world model can preserve small-object identity and predict precise placement into a shallow target region.

\section{Generated Rollout and Control Visualizations}
\label{app:rollout_visualization}

\begin{figure}[H]
    \centering
    \includegraphics[width=\textwidth]{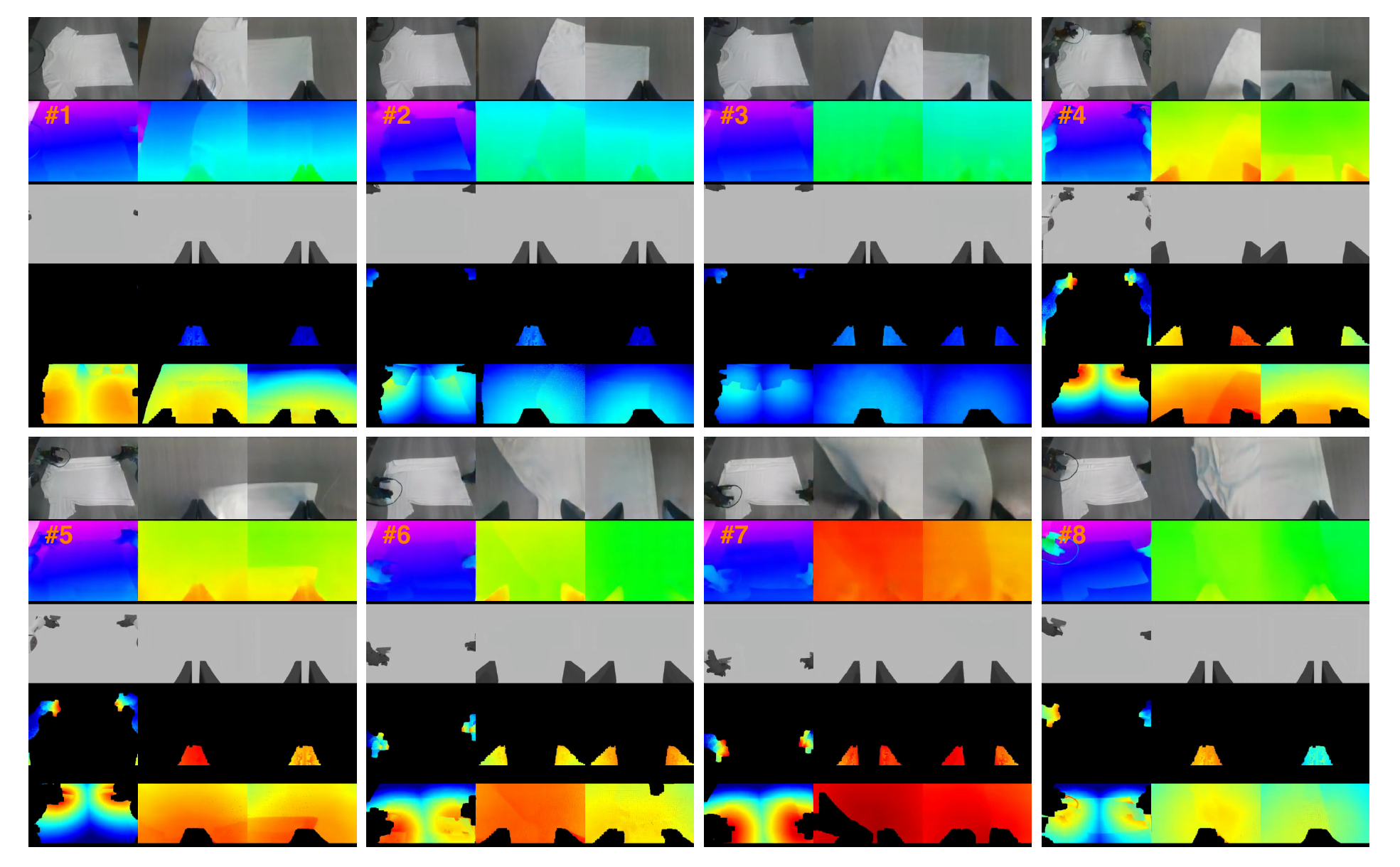}
    \caption{
    Long-horizon iMaC rollouts with paired video controls.
    For each task, generated RGB and depth are shown together with motion images and the two contact-image streams used during generation.
    }
    \label{fig:app_rollouts_controls}
\end{figure}

\begin{figure}[H]
    \centering
    \includegraphics[width=\textwidth]{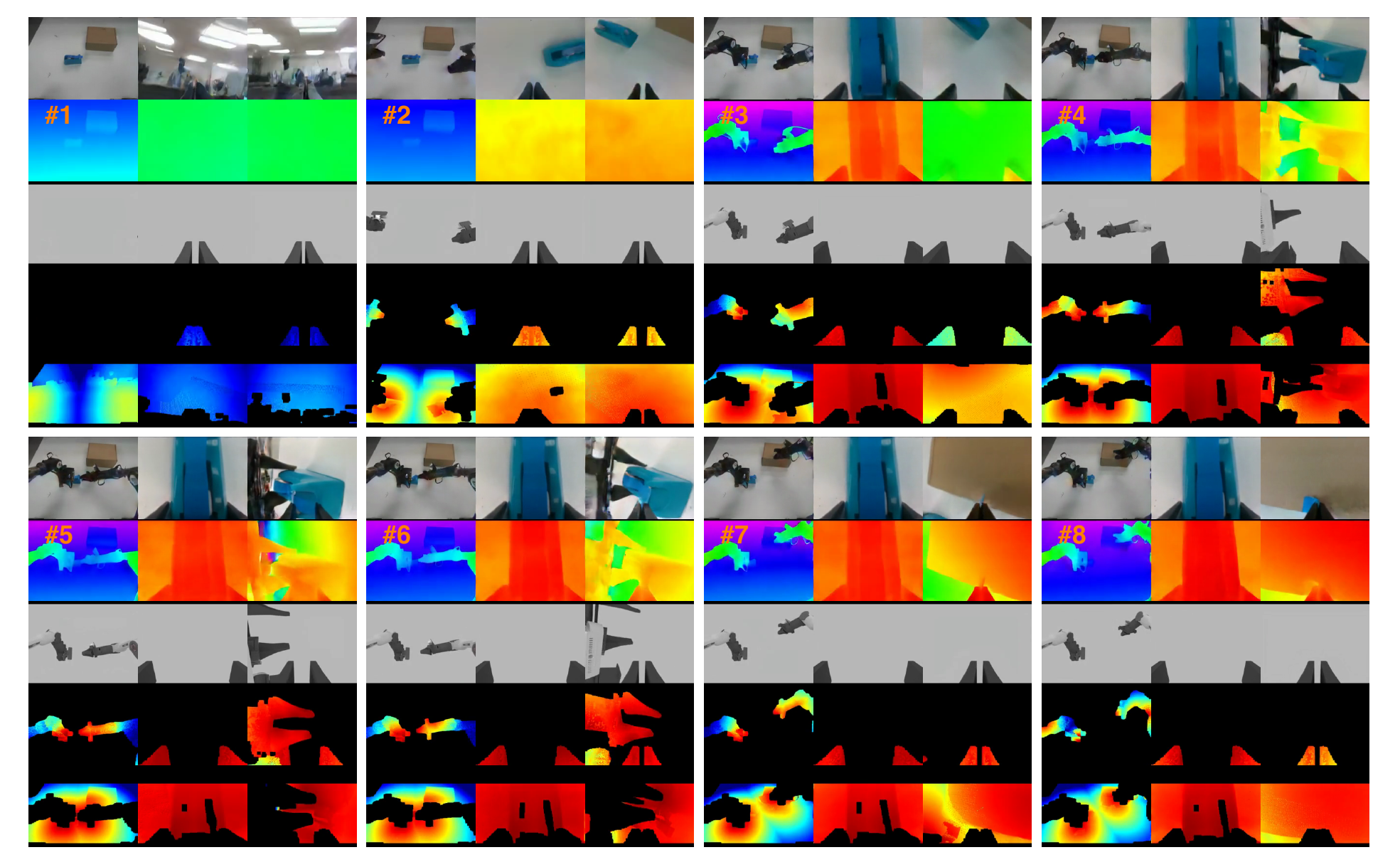}
    \caption{
    Additional iMaC rollouts with the same generated-video and video-control layout.
    The fixed layout shows how differences in generated scene motion are supported by the corresponding motion and contact images.
    }
    \label{fig:app_rollouts_controls_more}
\end{figure}

We include qualitative rollout visualizations in Figs.~\ref{fig:app_rollouts_controls} and~\ref{fig:app_rollouts_controls_more} to show how iMaC sustains long future-video generation under closed-loop chunk-wise rollout and how its generated videos align with the corresponding video controls.
Each example presents the reference observation, future RGB prediction, auxiliary depth prediction, URDF/FK-rendered motion images, and the two contact-image streams.
The first visualizations focus on normal generation behavior across tasks.
The generated RGB videos preserve the multi-view scene layout over long horizons, keep the robot motion consistent with the commanded trajectory, and produce object motion at the time and location suggested by the controls.
The auxiliary depth predictions remain spatially coherent enough to support subsequent pointcloud construction, even when RGB details become less sharp over later chunks.
Each figure uses the same row order so the reader can compare the generated video against the controls available to the model: motion images reveal the action-implied future robot body, while the scene-to-gripper and robot-to-scene contact images highlight where future gripper motion and robot geometry approach observed scene regions.

\begin{figure}[H]
    \centering
    \includegraphics[width=\textwidth]{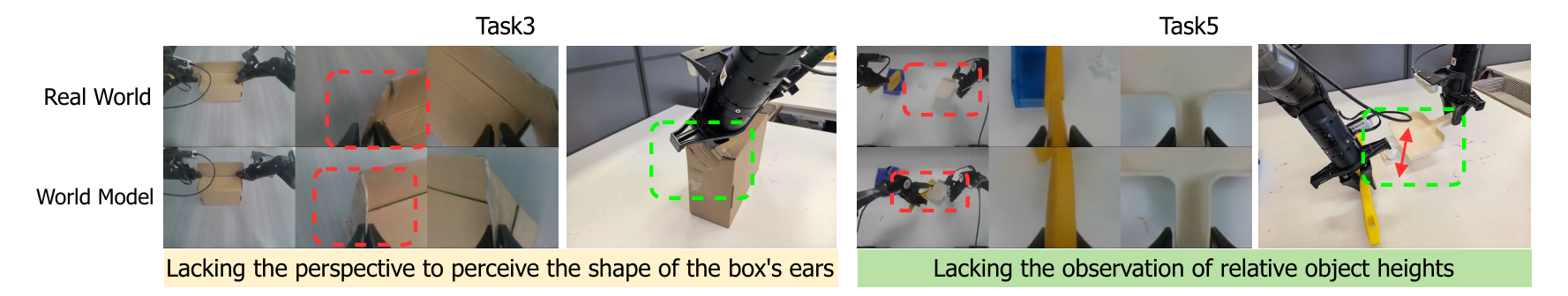}
    \caption{
    Focused failure-case visualization for missing task-relevant observations.
    Even with plausible generated videos and reasonable action-derived controls, the world model can mispredict scene evolution when all available views omit a physical relation that determines task success.
    }
    \label{fig:app_rollouts_controls_failures}
\end{figure}

\paragraph{Failure analysis.}
Long closed-loop video generation can fail in several ways.
Many failure cases reflect common limitations of current video models, including low visual fidelity, accumulated temporal error, and insufficiently accurate action following.
In our setting, however, we also observe a failure mode that is more directly tied to world-model-based policy evaluation: \emph{the model cannot reliably infer task-relevant physical relations that are missing from all available observations}.
Fig.~\ref{fig:app_rollouts_controls_failures} visualizes this boundary through two representative cases.
Task 3, ``stack the box ears,'' exposes this ambiguity in a box-ear insertion scenario.
The initial view can make the task appear as if the two side ears only need to be folded inward into the slot, but in the real scene the ears are long and can be blocked by the box side; the robot must first lift the ear so that its lower edge is above and aligned with the slot before inserting it.
The available views do not reliably reveal the height relation between the lower edge of the ear and the slot, so the model cannot determine how high the ear must be lifted to pass over the side wall.
As a result, it may generate an apparently plausible motion in which the ear is lifted and then lowered, while incorrectly predicting that the ear has entered the slot.
Task 5, ``sweep the trash into the dustpan using a broom,'' exhibits an analogous height ambiguity.
Whether the paper trash can be swept into the dustpan depends on the height of the dustpan entrance relative to the tabletop: if the entrance is flush with the table, the trash can be pushed inside, whereas a raised dustpan blocks the trash.
The three available views, including wrist-camera views, often do not provide a clean side observation of this height relation because the key local region can be occluded by the gripper, broom, or dustpan itself.
The world model may therefore generate plausible broom motion and trash displacement while mispredicting whether the trash can physically pass into the dustpan.
These examples suggest that, beyond improving video quality and control following, reliable learned real-world simulation also depends on camera coverage that captures the physical state variables that decide task success.

\end{document}